\documentclass[journal]{IEEEtran}
\usepackage{times,amsmath,epsfig}
\usepackage{epstopdf}
\usepackage{graphicx}
\usepackage{algorithm}
\usepackage{algpseudocode}							
\usepackage{cite}									
\usepackage[numbers,sort&compress]{natbib}			
\usepackage[numbers]{natbib}
\usepackage{tabularx,ragged2e,booktabs,caption}
\usepackage{subcaption}
\usepackage{mwe}
\usepackage{amsmath}
\usepackage{gensymb}
\usepackage[table]{xcolor}
\usepackage{booktabs,multirow,array}
\usepackage{xcolor,colortbl}
\usepackage[table]{xcolor}
\usepackage{adjustbox}
\usepackage{pbox}

\usepackage{hyperref}			
\hypersetup{
	colorlinks = false,
	linkbordercolor = {white},
}

\usepackage{xcolor}

\DeclareGraphicsExtensions{.pdf,.png,.jpg, .eps}

\ifCLASSINFOpdf
\else
\fi
\begin{document}
%

\title{Generation of High Dynamic Range Illumination from a Single Image 
for the Enhancement of Undesirably Illuminated Images}
%
%
%


\author{
	\IEEEauthorblockN{Jae Sung Park\IEEEauthorrefmark{1}\IEEEauthorrefmark{2}, Nam Ik Cho\IEEEauthorrefmark{1}} \\
	\IEEEauthorblockA{\IEEEauthorrefmark{1}INMC, Department of Electrical and Computer Engineering, Seoul National University, Seoul, South Korea \\
	jason79.park@ispl.snu.ac.kr, nicho@snu.ac.kr} \\
	\IEEEauthorblockA{\IEEEauthorrefmark{2}Visual Display Division, SAMSUNG Electronics Co. Ltd., Suwon, South Korea \\ }
}

\markboth{Journal of \LaTeX\ Class Files,~Vol.~14, No.~8, August~2015}%
{Shell \MakeLowercase{\textit{et al.}}: Bare Demo of IEEEtran.cls for IEEE Journals}

\maketitle

\begin{abstract}
This paper presents an algorithm that enhances undesirably illuminated images
by generating and fusing multi-level illuminations from a single image.
The input image is first decomposed into illumination and reflectance components by 
using an edge-preserving smoothing filter. Then the reflectance component is scaled up 
to improve the image details in bright areas. The illumination 
component is scaled up and down to generate several illumination images
that correspond to certain camera exposure values different from the
original. The virtual multi-exposure illuminations are blended
into an enhanced illumination, where we also propose a method to generate appropriate
weight maps for the tone fusion. 
Finally, an enhanced image is obtained
by multiplying the equalized illumination and enhanced reflectance.
Experiments show that the proposed algorithm produces visually pleasing
output and also yields comparable objective results to the conventional
enhancement methods, while requiring modest computational loads.
\end{abstract}


\IEEEpeerreviewmaketitle

\section{Introduction}
\label{s:Into}

\IEEEPARstart{M}{any} kinds of image enhancement and restoration
algorithms have been proposed for several decades 
to deal with so many kinds of image degradation \cite{gonzalez,baxes}. 
In this paper, we focus on
the degradation due to undesirable lighting conditions, for example 
low light or backlight situations \cite{ldr,gum,febm,npea,lime}. 
Images obtained under the low light conditions such as nighttime, 
underwater and medical images suffer from low contrast and low
signal to noise ratio. Also, the details in bright
or dark areas are often lost
in backlight and/or non-uniform 
lighting situations, because the dynamic
range of commercial cameras is not wide enough.

To improve the quality of degraded images captured in undesirable
illumination conditions, a variety of image enhancement methods have 
been proposed. Many of the existing methods are based on the histogram 
equalization \cite{ahe,bi-HE,clahe,ldr,hist-modi-framework}, which attempt to
expand the histogram of pixel values for boosting the
details in near-saturated areas. On the other hand, image filtering-based 
methods have also been proposed in 
\cite{wlsf,per-pix-exp,gum,fbf-hdr, auto-exp-correct, flash-photo},
which exploit edge-preserving filters for the manipulation of
decomposed image components or flash/non-flash image pairs.
In addition, image enhancement algorithms based on Retinex filtering 
have been proposed in \cite{cent-surr-retinex,febm,meylan,npea,framework-retinex}. 
According to Retinex theory, color perception of human visual system (HVS) 
has a strong correlation with the reflectance of objects and the characteristics of 
illumination \cite{cent-surr-retinex}. Hence, most of the Retinex-based 
approaches generally estimate the illumination component of
the input, and then remove it from the input to retrieve the 
reflectance component. Then the illumination components are separately 
processed to obtain the enhanced images \cite{febm,npea}. 
However, the estimated illumination images often look 
unreasonable when the input is acquired in the under-exposure 
or over-exposure conditions. To alleviate this problem, there
is also a more sophisticated method to estimate and/or refine 
the illuminations \cite{lime}.

Recently, single image high dynamic range (HDR) imaging methods have been proposed
for increasing the dynamic range of a single low dynamic range (LDR) image
\cite{febm, hdr-vexp-fusion, hdr-pexp-fusion}. 
The original HDR imaging technique is to capture multiple
images with different exposures, and produce larger bit depth
images by combining them or produce tone-mapped-like 
images by adding them with appropriate weights 
\cite{hdr-reinhard,hdr-exp-fusion}. The main part of HDR imaging 
is to obtain multi-exposure images by using time-division or
space-division multiple acquisition. The time-division
acquisition is also called the bracketing, which is implemented
in many of commercial cameras. This technique indeed widens the 
dynamic range for the static scene. However, it suffers from ghost artifacts
caused by moving objects and/or moving camera
while capturing the scene, and thus needs complicated
post-processing to remove the ghosts \cite{hdr-exp-fusion,an1,an2,an3}. 
On the other hand, the
space-division multiplexing methods using multiple sensors do not
have ghost artifact problem, but they need complicated registration and interpolation
process to match the view differences between the sensors 
\cite{hdr-sensor-array}. The single image HDR is to mimic this
process, i.e., it generates pseudo multi-exposure images from a single
input image and then follows the original HDR imaging method for
producing HDR images or tone-mapped-like exposure-fusion 
images for the LDR display.

In this paper, we propose a new singe image HDR imaging method
based on the Retinex filtering scheme, mainly for the purpose of
image enhancement under the undesirable illumination conditions.
The proposed algorithm first decomposes an input image into illumination 
and reflectance component by using an edge-preserving smoothing filter. 
Then, the reflectance component is scaled to improve the details 
in relatively bright areas. Also, we devise an algorithm that appropriately
scales up and down the illumination component, in order to generate 
several illumination images that correspond to certain camera exposure 
values different from the original. The conventional exposure fusion
method is applied to these illuminations to produce an equalized
illumination. Then the enhanced reflectance is multiplied with
the equalized illumination to produce the final enhanced image.
In this process, we also propose a
method to generate appropriate weight maps for keeping the contrast.

The rest of his paper is organized as follows. The overview of related 
works is presented in Section~\ref{s:R_works}. In Section~\ref{s:P_method}, 
the proposed illumination equalization method is discussed in detail. 
Experimental results and discussions are presented in 
Section~\ref{s:Experiments}. Finally, we draw our conclusions 
in Section~\ref{s:Conclusion}.

\section{Related Works}
\label{s:R_works}
The conventional methods for enhancing the undesirably illuminated images are
the histogram equalization (HE) and its variants such as adaptive histogram 
equalization (AHE) \cite{ahe}, contrast limited adaptive histogram 
equalization (CLAHE) \cite{clahe} and two-dimensional (2D) histogram
approach in \cite{cvc} named as contextual and variational contrast (CVC)
enhancement. Another approach is the single scale Retiex (SSR)
image enhancement method \cite{cent-surr-retinex}, which
finds the illumination-compensated result from the functional forms of
center/surround Retinex. Specifically,
the average intensity in the surround of a pixel corresponds to the
illumination and thus dividing this value from the original corresponds
to the illumination-compensated pixel value. 
However, due to the difficulty in selecting the scale parameter, multi-scale 
Retinex (MSR) that decides the output image by a weighted sum of SSR 
outputs with different scales is also developed, and multi-scale retinex with 
color restoration (MSRCR) was also proposed to remove the artifacts that
appear in MSR algorithm \cite{meylan,framework-retinex}.

Instead of histogram-based or Retinex filtering methods, a more intuitive 
approach to solving the illumination problem may be to appropriately adjust 
the illumination component, because the HVS is more sensitive to the
illumination changes than the color variations. There are 
illumination adjustment methods based on this principle 
\cite{febm, npea, lime}, and it is shown that they have the advantage
of preserving the naturalness while unveiling the details in dark shadow
areas. More precisely, Wang et. al. \cite{npea} proposed a 
naturalness preserving enhancement algorithm (NPEA) for improving 
the non-uniformly illuminated images. Based on the Retinex approach, an 
input image is decomposed into reflectance and illumination component 
by using a bright-pass filter. Then the estimated illumination component
is adjusted by a bi-log transformation, while trying to preserve 
the naturalness by observing the measure of lightness-order error. 
In the case of low-light image enhancement (LIME) method proposed in 
\cite{lime}, they mainly focused on the exact estimation of illumination 
component. In detail, initial illumination of each pixel is estimated by 
individually finding the maximum values in three color channels 
(Red, Green and Blue). Then, a structure-aware smoothing model that 
is developed to refine the initial illumination component is applied to 
obtain the sophisticated illumination. For the reflectance component, 
a denoising technique is employed to suppress possible noise in dark 
areas. Finally, they reproduced an improved image by combining the 
adjusted illumination and the denoised reflectance.
In fusion-based image enhancement method (FbEM) \cite{febm}, they first 
estimated the illumination component of input image by using the
brightest channel prior, which is inspired from the dark channel 
prior (DCP) used for the dehazing algorithms \cite{ex:DCP}.
Then, two different versions of 
illuminations, luminance-improved and contrast-enhanced illumination 
components, are derived by using an adaptive histogram equalization and 
a sigmoid function. Then the appropriate weights are designed for these 
illumination components, and the weighted illuminations are blended to 
be a final bright and contrast enhanced illumination. Finally, they 
reproduced an improved image by combining the adjusted 
illumination with the reflectance.

\begin {figure} 
\centering
\includegraphics[width=90mm]{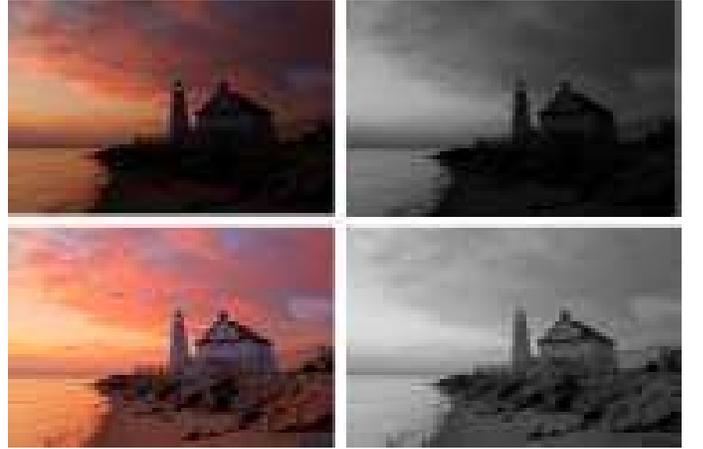}
\caption{A non-uniformly illuminated image and its illumination equalized 
	result: (clockwise from top left) an input image 'Lighthouse,' 
	estimated illumination, equalized illumination by adding the pseudo multi-exposure
illuminations, and enhanced image using the equalized illumination.}
\label{Figure0_Motivation}
\end {figure}

\section{Proposed Enhancement Algorithm}
\label{s:P_method}
\subsection{Motivation and Overview}
The dynamic range of most commercial cameras is limited so that
they have problem with wide dynamic range scenes or
with the very low light conditions. Specifically, the images often 
lose details in some regions when they are taken under undesirable 
illumination conditions such as backlight, non-uniform light or 
extremely low light environment. In this paper, we attempt to 
enhance such images based on the Retinex filtering, i.e., we 
estimate illumination and reflectance of the image and process 
them separately. 

In general, it is difficult to obtain the accurate 
illumination by using the conventional filter-based estimation
methods. Since the illumination is estimated as the 
weighted average of neighboring pixels, it is affected by 
the reflectance and thus does not appear uniform even in 
uniformly illuminated areas as shown in Fig.~\ref{Figure0_Motivation}.
The top-left figure is the input image, where the shore area is 
supposed to have uniform illumination change. 
However, the estimate of illumination (top-right figure)
in this area does not appear uniform because it 
is affected by the reflectance values of ocean and sand. 
Hence we attempt to obtain 
more equalized illumination via the fusion
of pseudo multi-exposure illuminations as shown in 
the bottom-right figure, and 
multiply it with the reflectance to obtain the equalized 
output in the bottom-left of Fig.~\ref{Figure0_Motivation}.

Our algorithm is basically based on the single HDR approach.
By increasing and decreasing the 
illumination to several levels, we generate several
illumination images which correspond to different camera exposure values.
Multiplication of these illuminations to the reflectance produces pseudo
multi-exposure images, and then we can finally obtain the
enhanced image by using a multi-exposures fusion algorithm.

In summary, our contribution is to design a simple and effective 
method for enhancing the estimated reflectance and  
illumination, and also to further enhance the overall image 
by generating pseudo multi-exposure illuminations and combining them 
in the manner of exposure fusion. The most important and 
sophisticated process in our algorithm is to generate multiple 
illuminations that correspond 
to the multiple LDR images of different exposures for HDR imaging, 
without any prior information (capture device, exposure 
conditions etc.) on the input images.

\begin {figure}
\centering
\includegraphics[width=70mm]{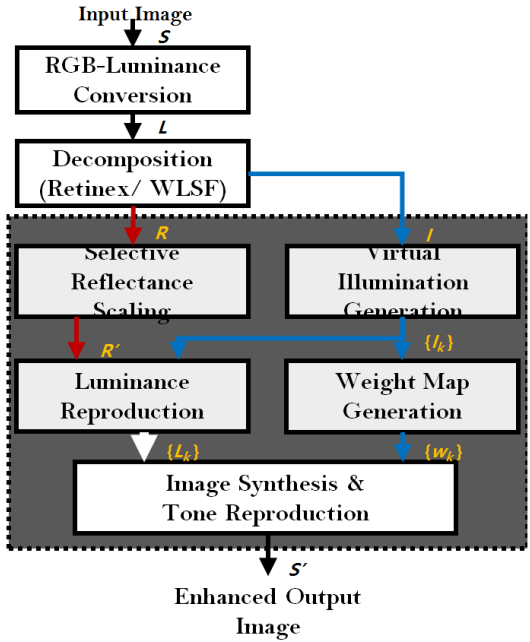}
\caption{Block diagram of the proposed method.}
\label{Figure1_block_diagram}
\end {figure}

The block diagram of our method is depicted in 
Fig. \ref{Figure1_block_diagram} and the contributions  
are indicated in the dark gray box. The proposed 
algorithm is consisted of three steps. The first step is to 
extract the luminance information ($L$) from an input RGB image ($S$) 
by the RGB to Lab transform, and then to decompose $L$ into illumination ($I$) and 
reflectance component ($R$). We employ the 
weighted least square filter (WLSF) instead of Gaussian
filter that is frequently used in the conventional methods
in order to prevent the halo artifacts. The second step is to 
adjust the reflectance and illumination component, and
then to generate virtual illuminations that correspond
to several different exposure values. For the adjustment
of reflectance, we propose a selective reflectance scaling 
(SRS) method which enhances the details in relatively bright
areas, which contributes to contrast enhancement. 
Simultaneously, we equalize the illumination and then
generate several illuminations by developing a
virtual illumination generation (VIG) method.
Specifically, our VIG generates a set of multiple
illuminations ($\{I_k\}$) which correspond to the ones in 
multi-exposure HDR imaging techniques. Then, a set of new luminance 
images ($\{L_k\}$) is reconstructed by combining the 
detail-improved reflectance ($R'$) and the set of illumination 
components ($\{I_k\}$). We also design appropriate weights 
($\{w_k\}$) that contribute to preserve the image details and
to enhance brightness in every image area. Finally,
the luminance images are fused to generate an enhanced 
one. By adopting tone reproduction process for the final 
luminance image, we obtain a brightness and detail-enhanced image.

\subsection{Image Decomposition}
\label{ss:decomp}
The proposed enhancement algorithm starts with the Retinex-based 
illumination and reflectance decomposition. The luminance is
first obtained from the input RGB image, and the reflectance 
information is obtained by the difference between the 
input luminance and the estimated illumination as
\begin{equation}
R\left(i,j\right) = log(L\left(i,j\right))-log(I\left(i,j\right))
\label{eq1: Retinex}
\end{equation}
where ($i$,$j$) is the index for pixel position, 
$R$ corresponds to the reflectance information,  
$L$ is the luminance, and $I=(L*G(\cdot))$ is the
estimate of illumination obtained by the filtering of $L$ 
with a filter $G$ which is usually a normalized Gaussian function
in the conventional works. However, 
it was reported that the Gaussian kernel generally makes halo 
artifacts near the border of bright background and dark 
foreground \cite{pa:retinex-wlsf}. In general, the
Gaussian filter needs a very large support such that the 
illumination estimation
is less affected by reflectance values. However, this 
produces wrong estimates around the strong edges, which
results in halo artifacts.
Hence, we adopt the weighted least square filter (WLSF) instead 
of Gaussian to reduce the halo artifacts. The WLSF is a kind of 
edge-preserving and smoothing filters, where the filtering result
is obtained by minimizing the following energy function:
\begin{multline}
E = \sum_p\left(\left(u_p-L_p\right)^2\right) + \\ \lambda \left( a_{x,p}\left(L\right)\left(\frac{\partial u}{\partial x}\right)^2_p + a_{y,p}\left(L\right)\left(\frac{\partial u}{\partial y}\right)^2_p \right)
\label{eq1: WLSF}
\end{multline}
where $u$ is the output of the filtering, $\lambda$  
is the balancing parameter between the first and the second terms, and the subscript 
$p$ denotes the pixel position. In the above energy function, the first 
term $\left(u_p-L_p\right)^2$ is to prevent the output $u$ from being 
too much deviated from the input $L$, and the second is a smoothness 
term that suppresses textures by minimizing the partial derivatives 
$\left({\partial u}\right)/\left({\partial x}\right)_p$ 
and $\left({\partial u}\right)/\left({\partial y}\right)_p$. 
They can be further strengthened or weakened according to the strength 
of input texture by defining the
weights $a_{x,p}\left(L\right)$ and $a_{y,p}\left(L\right)$ as
\begin{equation}
\begin{split}
a_{x,p}\left(L\right) = \left(\left|\frac{\partial l}{\partial x}
\left(p\right)\right|^\alpha + \epsilon \right)^{-1} \\
a_{y,p}\left(L\right) = \left(\left|\frac{\partial l}{\partial y}
\left(p\right)\right|^\alpha + \epsilon \right)^{-1}
\end{split}
\label{eq1: WLSF_weights}
\end{equation}
where the $\alpha$ is a control parameter for the sensitivity to 
the gradients of input $L$, and $l$ is the log-luminance  
defined as $l$ = $log(L + \epsilon)$ with a very 
small constant $\epsilon$ to prevent division by zero. 
The output of this WLSF is 
used as our initial illumination component ($I$).

\begin {figure} 
\centering
\includegraphics[width=70mm]{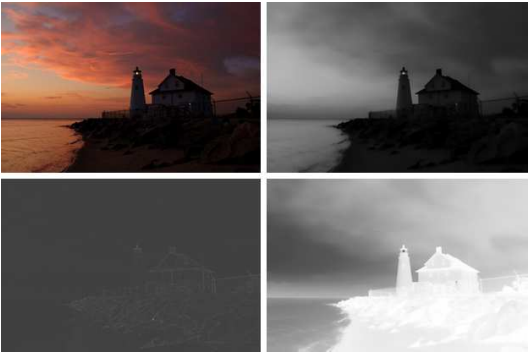}
\caption{Image decomposition: (clockwise from top left) an LDR input 
	image of single exposure, estimated illumination component 
	(corresponding to BM), inverse of BM (corresponding to DM) 
	and reflectance component (adjusted for visualization).}
\label{Figure2_Retinex}
\end {figure}

However, like any other filter based estimation methods,
the WLSF also underestimates the illumination in dark shadow areas
or dark colored objects and thus the estimated illumination appears 
very small in such areas. Considering that increasing the
brightness and contrast of dark areas is the basics of many image
enhancement and HDR imaging techniques, we also increase the
illumination in dark areas as long as the color is not too much distorted.
In this process, the illumination in bright areas should not be affected
too much, and thus we propose to adjust the illuminations of
bright and dark areas separately. For this locally different 
adjustment strategy,
we define two maps that represent the brightness/darkness
of the illumination intensities. 
Specifically, the first
map is named as the brightness map (BM) 
which is the image of normalized illumination intensity,
and the second is the darkness map (DM) which
contains the inverse of illumination at each pixel position.
As an example, Fig. \ref{Figure2_Retinex} shows an input image, 
BM, DM and reflectance components. 
We control the illumination changes considering these
two intensity maps, which will be explained in Section \ref{ss:VIG}. 

\subsection{Selective Reflectance Scaling (SRS)}
\label{ss:SRS}
\begin {figure} 
\centering
\includegraphics[width=90mm]{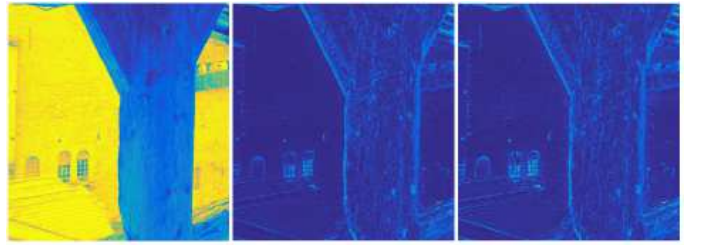}
\caption{Scaling the reflectance component by SRS: 
	(from left to right) a cropped area of the image shown 
	in Fig. \ref{Figure2_Retinex}, original reflectance of this area,
	and the enhanced reflectance, all represented in false color
	(blue: dark, yellow: bright).}
\label{Figure3_SRS}
\end {figure}

For enhancing the details in dark areas, illumination is going 
to be adjusted
as will be described in the next subsection. However, in the case of 
bright areas, increasing the illumination may saturate the pixel values. 
Also, sharpness or details cannot be sufficiently enhanced in 
relatively well-illuminated areas since our illumination equalization 
method is not pixel-wise processing. For these reasons, instead of
adjusting the illumination, we stretch the reflectance
at bright areas where the illumination component is larger than a certain
threshold. Precisely, the reflectance is modified according to 
\begin{align}
R'(i,j)=\begin{cases}
R(i,j)\left(\frac{I\left(i,j\right)}{m_{I}}\right)^{\gamma_{R}}, 
& I(i,j) > m_I \\
R\left(i,j\right), & \mbox{otherwise} \\
\end{cases}
\label{eq2: SRS}
\end{align}
where $m_{I}$ is the average value of the estimated illumination and the 
parameter $\gamma_{R}$ is the gamma value which is set to 0.5 in this paper. 
Hence when the illumination of a pixel $I(i,j)$ is larger than the mean 
$m_{I}$, then it is considered a bright pixel and the reflectance is 
increased by the amount of
$\sqrt{I(i,j)/m_I}$. As a result, the details in bright areas 
become clearer than before. As an example, 
Fig. \ref{Figure3_SRS} 
shows the effect of the proposed algorithm. The first 
image is a crop of the image
in Fig. \ref{Figure2_Retinex}, where the illumination of the pixels is
shown in false color representation from yellow (bright) to blue (dark). 
The second is the reflectance value also in false color representation
and the third is the result of applying our SRS in eq.~\ref{eq2: SRS}.
It can be observed that the texture (especially at the bright background)
becomes apparent as a result of SRS.
To validate the effect of SRS objectively, 
the 'discrete entropy (DE)' \cite{pa:de} is
measured for the second and third images in Fig.~\ref{Figure2_Retinex},
resulting in 2.30 and 2.42 respectively. The DE is devised to measure the 
total amount of information in the image and thus we can say that 
more information is unveiled by the SRS.
The reason why we do not apply our SRS algorithm to dark areas is
that it can amplify the noise in dark areas. Too much increase
of brightness in dark areas can also reduce the
overall contrast, which often makes the image seem unnatural.
Hence the dark areas are improved by the illumination fusion
method as follows.

\subsection{Virtual Illumination Generation (VIG)}
\label{ss:VIG}
With a single input luminance $L$ with unknown exposure value (EV), we 
design a simple and effective algorithm that generates multiple 
virtual illuminations. Specifically, we design a ``scale function''
that outputs a scale factor for the given virtual EV. Then the scale
factor is multiplied to $L$ to brighten or darken the original
illumination according to the given EV.

\begin {figure} 
\centering
\includegraphics[width=90mm]{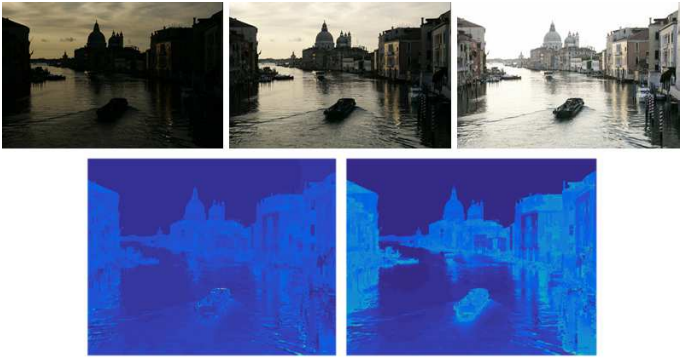}
\caption{Muli-exposure images and the relationship between the estimated illuminations: 
	(top) under-, standard- and over-exposure LDR images of the 'Grandcanal' with 
	exposure times of 1/750, 1/180 and 1/45, (bottom left) ratio of illumination of 
	standard-exposure image to that of under-exposure image, (bottom right) 
	ratio of illumination of over-exposure image to that of under-exposure image.}
\label{Figure_VIG_int}
\end {figure}

For this, we first observe 
the illumination change in a set of real 
multi-exposure images acquired by 'bracketing.' Specifically, 
consider a set of multi-exposure LDR images shown in 
Fig.~\ref{Figure_VIG_int} (top three),
which are under-, standard- and over-exposure ones from left to right  
\cite{pa:hdrsoft-web}. We denote the images as 
$S_u$, $S_s$ and $S_o$ and their illuminations
$I_u$, $I_s$ and $I_o$ respectively. 
To investigate the tendency of illumination changes for these images, 
the ratios of illuminations are also shown in 
Fig.~\ref{Figure_VIG_int}.
The bottom-left image shows the ratio of 
standard-exposure to under-exposure, 
i.e, ($I_s/I_u$), and the bottom-right is the ratio of 
over-exposure to under-exposure ($I_o/I_u$). 
The figures clearly show that the ratio in shadow area is 
much larger than that in 
bright area. We take this observation into consideration, i.e., 
we adjust the illumination more in the dark regions than 
in the bright areas. 

Also, we need to consider another aspect of
camera imaging system that the scene radiance is nonlinearly mapped
to the pixel values according to the sensor properties and other camera 
electronics. A global illumination adjustment method considering this 
property was proposed in \cite{pa:adapt-exp-correct}, 
by designing a function that finds a global scale factor 
for the given EV.
However, we do not take this approach because we consider more 
general case without EV information, and also because we 
are locally adjusting the illumination. 
Since we do not have EV for the given input, we instead define
\textit{virtual exposure value} (vEV), which is the average of normalized
illumination in the range of [0,1]. We devise a function
that maps a vEV to a scale factor, where the range of scale factor
is adjusted according to the overall illumination. The
illumination is globally adjusted according to the scale
factor, and then locally adjusted depending on the local
brightness.  
 
\begin{figure}
	\centering
	\begin{subfigure}[b]{0.3\textwidth}
		\includegraphics[width=1\linewidth]{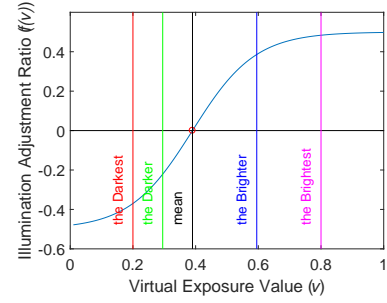}
		\caption{}
		\label{Figure5_VIG_Graph_low}
	\end{subfigure}	
	\begin{subfigure}[b]{0.3\textwidth}
		\includegraphics[width=1\linewidth]{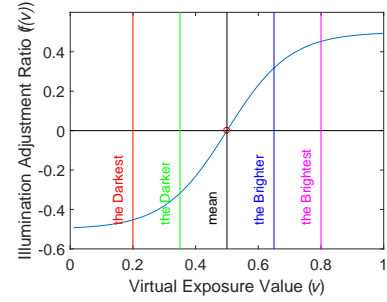}
		\caption{}
		\label{Figure5_VIG_Graph_avg}
	\end{subfigure}
	\begin{subfigure}[b]{0.3\textwidth}
		\includegraphics[width=1\linewidth]{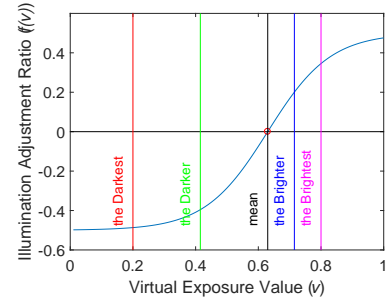}
		\caption{}
		\label{Figure5_VIG_Graph_high}
	\end{subfigure}
	\caption {The scale function for illumination adjustment.
Examples for three cases are shown, (a) for relatively small average value $m_I<0.5$, (b) for $m_I=0.5$
and (c) for relatively large average value $m_I>0.5$. 
Four different vEVs (from darkest to
brightest) are indicated in each graph
for generating four different virtual illuminations.}
	\label{Figure5_VIG_Graph}
\end{figure}

Our scale function is in the form of sigmoid, which
has been commonly used for the tone-mapping
and/or illumination compensation in the conventional works.
Specifically, it is defined as
\begin{equation}
f(v)= r \left(\frac{1}{1+e^{-\sigma_s \left(v-m_I\right)}}-\frac{1}{2} \right)
\label{eq: VIG}
\end{equation}
where $v$ is the vEV in the range 
of [0,1], $m_I$ is the average of the normalized 
illumination of original image, $\sigma_s$ is the smoothness 
factor of the sigmoid curve which is just set to 1 in all of our 
experiments, and $r$ is the adaptive amplitude which controls
the range of the scale function.
Fig.~\ref{Figure5_VIG_Graph} shows some examples of this function
for the cases of $m_I < 0.5$ (original image is rather dark on average), $m_I=0.5$ (well-balanced input image), and
$m_I > 0.5$ (bright image in overall region). 
It can be seen that the inflection point is set to
$m_I$ and the value of the function is 0 at this point.
We can also see that the
range of the function is larger when $m_I$ is small
in order to scale the illumination more for the darker images.
The range of the function is controlled by $r$, 
which needs to be large when $m_I$ is small and vice versa. This range
control is easily implemented by defining it as
\begin{equation}
r = \left(1-\frac{m_{I}}{M}\right) .
\label{eq:VIG_param}
\end{equation}
where $M$ is the maximum value of the illumination.  

In the experiments, we generate four different illuminations
that correspond to different vEVs, so that
we have five illuminations including the original (vEV$=m_I$).
Specifically, we define five $v_k$ ($k=1,2,3,4,5$) where $v_3=m_I$,
$v_1=0.2$, $v_5=0.8$, $v2=(v_1+v_2)/2$, and 
$v4=(v_3+v_5)/2$ as illustrated in Fig.~\ref{Figure5_VIG_Graph_avg}.
In the figure, it can be seen that $v_1$ to $v_5$ correspond to
the darkest, darker, original, brighter, brightest images.
If $m_I$ is less than 0.2 or larger than 0.8,
then it means that most part of original image is saturated and thus we
give up the enhancement. In all the experiments, there was actually no
case that $m_I$ is out of this range. 

The global adjustment is to multiply $f(v_k)$ to the original
illumination and add it to the original, i.e.,  
\begin{equation}
I_{k}(i,j) =(1+f(v_k))I(i,j).
 \label{eq:global}
\end{equation}
As stated previously, we add local adjustment to the above equation
for boosting the illumination change at dark areas. This is
implemented as  
\begin{equation}
I_{k}(i,j) =(1+f(v_k))(I(i,j)+f(v_k)\breve{I}(i,j))	
\label{eq:VIG_final}
\end{equation}
where $\breve{I}$ is the inverse of the normalized initial illumination 
$I$, i.e., DM which is defined in Section~\ref{ss:decomp} 
and shown in Fig.~\ref{Figure2_Retinex}. Since the DM is large
at the dark areas, the eq.~\ref{eq:VIG_final} can enlarge
the change of illumination more than the eq.~\ref{eq:global}
at the dark areas.

\begin{figure}
	\centering
	\begin{subfigure}[b]{0.5\textwidth}
		\includegraphics[width=1\linewidth]{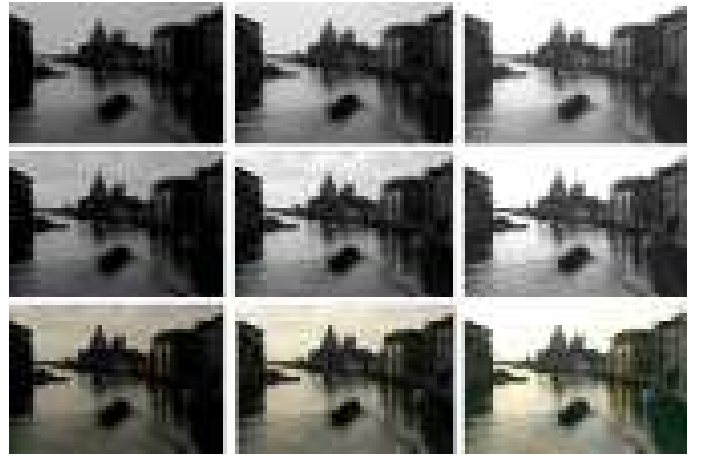}
		\caption{}
		\label{Figure6_VIG_Spl1_canal}
	\end{subfigure}	
	\begin{subfigure}[b]{0.5\textwidth}
		\includegraphics[width=1\linewidth]{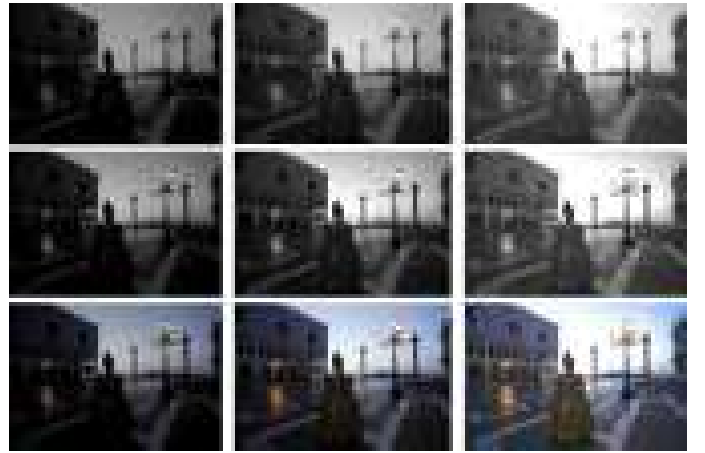}
		\caption{}
		\label{Figure6_VIG_Spl1_mask}
	\end{subfigure}
	\caption {Generated virtual illuminations ($L_k$) and 
		tone-reproduced images ($S_k$) from the virtual illuminations ($I_k$).
		The first row shows $I_k$, the second row $L_k$ and the third
		$S_k$ for $k=1, 3, 5$ for the images of (a) Grandcanal and (b) Mask.}
	\label{Figure6_VIG_Samples}
\end{figure}


\subsection{Tone Reproduction}
\label{ss:T_repro}
The next step is to generate pseudo multi-exposure luminances
from the enhanced reflectance $R'$ and the $I_k$. For this
we first obtain the luminance image for each vEV as:
\begin{equation}
L_k\left(i,j\right) =  \exp \left({R'\left(i,j\right)}\right)I_k\left(i,j\right).
\label{eq: Lum_repro}
\end{equation}
The generated luminance images are shown in Fig.\ref{Figure6_VIG_Samples}
for three vEVs. The first row shows the normalized illumination
$I_k$ for $k=1,3,5$, and
the second row shows the corresponding luminance images $L_k$. 
Additionally, differently illuminated RGB images $S_k$ are shown 
in the third row. The $S_k$ is obtained by using a conventional tone 
reproduction method \cite{pa:kim-tone-repro}: 
\begin{equation}
S_k = \left[ \begin{array}{c} Red' \\ Green' \\ Blue' \end{array} \right]_k 
= \left[ \begin{array}{c} L_k\left(\frac{Red}{L}\right)^{\gamma_{t}} \\ L_k\left(\frac{Green}{L}\right)^{\gamma_{t}} \\ L_k\left(\frac{Blue}{L}\right)^{\gamma_{t}} \end{array} \right]
\label{eq: tone_repro}
\end{equation}
where $\gamma_{t}$ 
denotes the gamma correction coefficient in the range of [0,1]. 
In this paper, $\gamma_{t}$ is just set to 1. As shown in the third 
rows in each image set in Fig. \ref{Figure6_VIG_Samples}, they 
appear similar to multi-exposure LDR images obtained by bracketing.

\subsection{Weight Rules for Luminance Synthesis}
\label{ss:W_rule}
\begin {figure} 
\centering
\includegraphics[width=90mm]{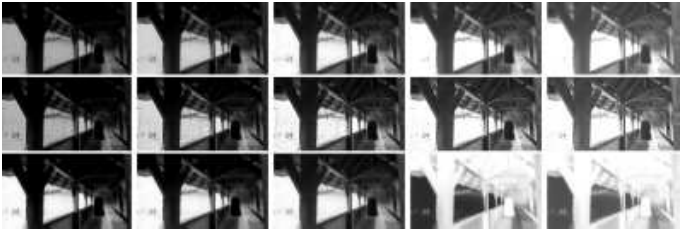}
\caption{Weight maps of differently exposed luminance images.
	First row: pseudo illuminations $I_k$, $k=1,2,\cdots,5$.
	Second row: reproduced luminance images $L_k$. Third: weight
	maps $w_k$.}
\label{Figure7_Blending}
\end {figure}

To blend the pseudo multi-exposure luminances as an illumination-equalized 
single luminance, we design an appropriate weighting rule which
focuses on unveiling the details in dark areas and simultaneously 
maximizing the overall contrast. 
For this, the bright areas need to be stretched in the
case of under-exposed images and conversely dark areas
be stretched in the case of over-exposed ones.
Precisely, for the under-exposed image, we take all the details 
in the brightest areas by setting the weights in these areas
larger than others. 
This is simply implemented by setting the weight map as the
normalized illumination values.
Conversely, for the over-exposed luminance image, we take the details 
in the dark areas. Hence, we let the weighting values  
in the dark area be larger than those in other regions. This is 
achieved by setting the weight map as the inverse of normalized 
illumination. In summary, the weight maps are represented as 
\begin{align}
w_k=\begin{cases}
\bar{I_k}, & \text{$k$=1, 2, 3}.\\
{\bar{I_k}}^{-1}, & \text{$k$=4, 5}.\\
\end{cases}
\label{eq:blending_weight}
\end{align}
where ${\bar{I_k}}$ is the normalized illumination,
and ${\bar{I_k}}^{-1}$ means that each value in
the illumination is inversed. The illumination images,
luminance, and their weight maps are visualized in 
Fig.~\ref{Figure7_Blending}. 
There are two advantages in our weighting strategy. First, since the 
estimated illuminations ($I_k$) inherit the contrast of the 
input, the weight maps and the adjusted illuminations 
approximately succeed the image contrast. Second, our weight
map is simple to obtain compared to the ones in the conventional
exposure fusion methods \cite{hdr-reinhard,hdr-exp-fusion}, 
which need complicated measures
such as contrast, saturation and exposedness of pixels and/or regions. 

Finally, we can obtain an illumination-equalized luminance  
by applying the conventional weighted-averaging as 
\begin{equation}
L'\left(i,j\right) = \frac{\sum_{k=1}^{N}{L_k\left(i,j\right)w_k\left(i,j\right)}}{\sum_{k=1}^{N}{w_k\left(i,j\right)}},
\label{eq:image_blending}
\end{equation}
where $L'$ represents the final luminance image
and $w_k$ is the
blending weight map for the $k$-th luminance image. The final
enhanced image in RGB color space can be also obtained by using the eq.~\ref{eq: tone_repro} with the final luminance image and color information of the original image.

\section{Experiments and Discussions}
\label{s:Experiments}
Experiments are performed with the images from the other papers
and MMSPG dataset \cite{ex:mmspg} which are captured in various 
illumination conditions such as backlight, low light and non-uniform
light environments. 

\subsection{Comparison with other methods}
The proposed method is compared with the conventional 
histogram-based methods (CLAHE \cite{clahe} and CVC \cite{cvc}), 
a retinex-based algorithm (MSR) \cite{meylan},
and illumination adjustment methods such as NPEA, FbEM and LIME
\cite{npea,febm,lime}. The parameters for these methods are
set the same as the original work when they are clearly
specified in the paper. Specifically, all the parameters 
for the CLAHE are set to be the default values. 
The parameters for CVC are set as $\alpha= 1/3$, $\gamma= 1/3$, 
and $\beta=2$, and $7 \times 7$ block size is used. For MSR, the weights 
for three SSR results are set the same, and the patch sizes of 
three Gaussian functions are set to 15, 80, and 250 respectively. 
For NPEA, FbEM and LIME, the encrypted Matlab codes provided by the 
authors are used. 

\subsubsection{Subjective comparisons}
\begin {figure*} 
\centering
\includegraphics[width=150mm]{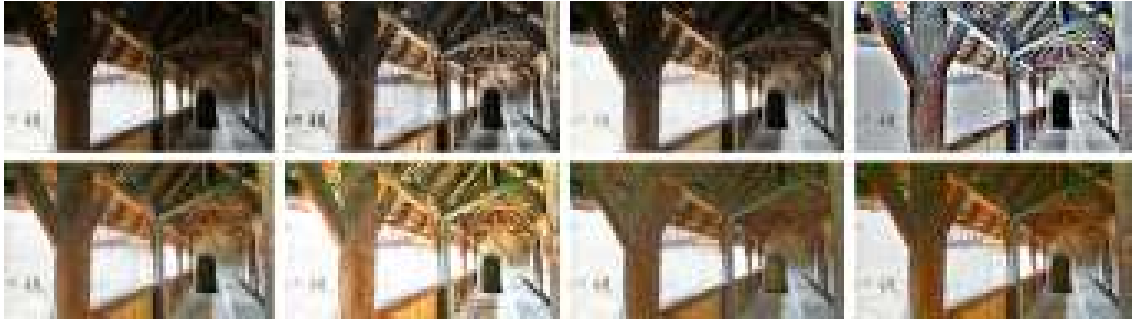}
\caption{Results for the image 'Wooden ceiling' labled as C23 in MMSPG dataset.
	From top left to bottom right: input image, result of CLAHE, CVC, MSR, FbEM, 
	LIME, NPEA and proposed method.}
\label{Figure8_Exp_Rst_BL1}
\end {figure*}

\begin {figure*} 
\centering
\includegraphics[width=150mm]{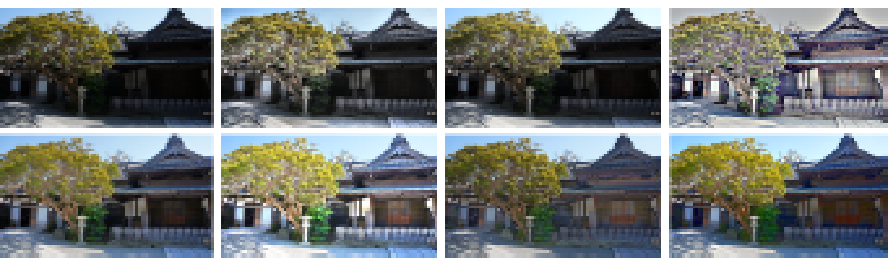}
\caption{Results for the image 'Tile-roofed house' labled as C11 in 
	MMSPG dataset.
	From top left to bottom right: input image, result of CLAHE, CVC, MSR, FbEM, 
	LIME, NPEA and proposed method.}
\label{Figure9_Exp_Rst_BL2}
\end {figure*}

First, the results for backlight images, which contain very bright and 
dark areas simultaneously, are presented in Figs.~\ref{Figure8_Exp_Rst_BL1} 
and \ref{Figure9_Exp_Rst_BL2}. As can be observed, CLAHE does not properly
improve the brightness especially in the dark areas, and the CVC produces 
somewhat blurred and unnatural results. Although the MSR enhances details 
in dark areas clearly, there are some color distortions. 
On the other hand, the FbEM, LIME, NPEA and the proposed method 
show enhanced details in the shaded areas. 
In the case of LIME, however, it seems that the brightness
is too much amplified. 

\begin {figure*} 
\centering
\includegraphics[width=150mm]{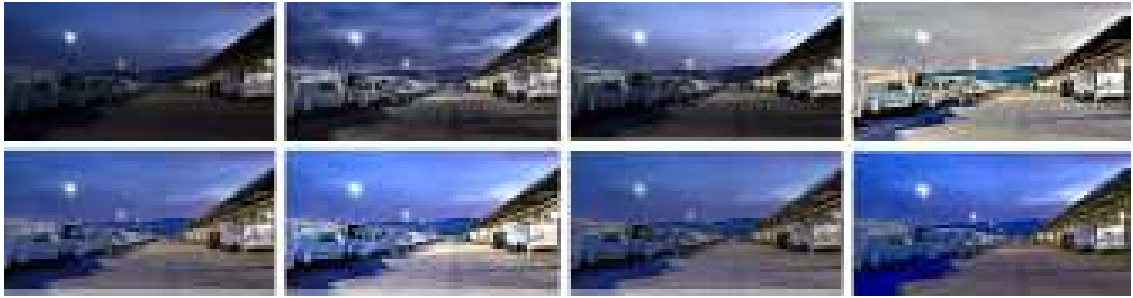}
\caption{Results for the input image, 'Daybreak'. (from top left to bottom right) Input image, result of CLAHE, CVC, MSR, FbEM, LIME, NPEA and proposed method.}
\label{Figure10_Exp_Rst_Low1}
\end {figure*}

\begin {figure*} 
\centering
\includegraphics[width=130mm]{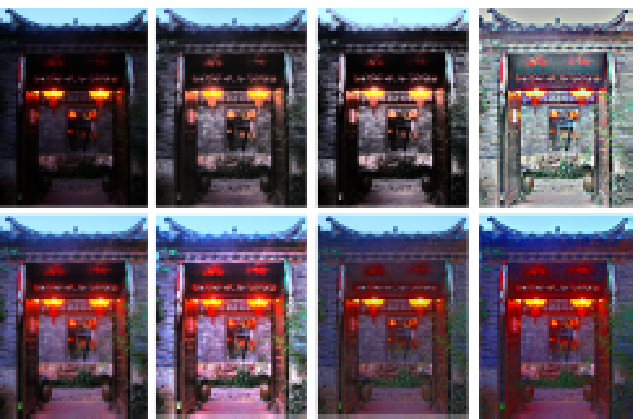}
\caption{Results for the input image, 'Entrance'. (from top left to bottom right) Input image, result of CLAHE, CVC, MSR, FbEM, LIME, NPEA and proposed method.}
\label{Figure11_Exp_Rst_Low2}
\end {figure*}

Figs.~\ref{Figure10_Exp_Rst_Low1} and \ref{Figure11_Exp_Rst_Low2} show 
the results for the images captured in the low light conditions. 
Similar to the backlight case, CLAHE and CVC fail to unveil 
the details in all areas. The MSR enhances details in all areas but the 
results look unnatural. The LIME over-enhances the image 
such that the results give the impression that the photo is
taken at daytime while it was actually taken at dawn or night. 
The FbEM and the 
proposed algorithm generate similar result, but the proposed method shows
more contrasted image and hence 
shows more vivid texture. 

\begin {figure*} 
\centering
\includegraphics[width=150mm]{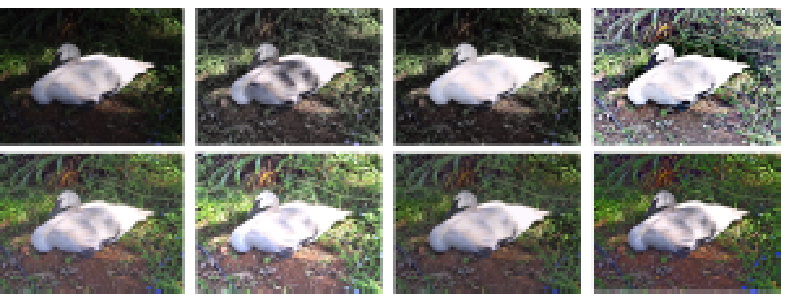}
\caption{Results for the input image, 'Duck'. (from top left to bottom right) Input image, result of CLAHE, CVC, MSR, FbEM, LIME, NPEA and proposed method.}
\label{Figure12_Exp_Rst_NonUni}
\end {figure*}

Results for the images taken under the non-uniform illumination condition 
are presented in Fig. \ref{Figure12_Exp_Rst_NonUni}. The non-uniform 
illumination means that there are low and highly illuminated regions within
small areas. It is clear that the CLAHE generates a contrast-enhanced result, 
but it does not look natural as the texture on the duck becomes too dark. 
The CVC and MSR over-stretch the contrast and hence the results lose the 
details in bright areas. On the other hand, the FbEM, LIME, 
NPEA and the proposed method show quite pleasing results. 
Each result has its own pros and cons and the preference may
be different from person to person. But, like the above cases, the LIME
stretches bright areas too much such that the texture on the duck's
feather is lost.

\subsubsection{Objective comparisons}
Since our HVS and its perception mechanism are so complicated, 
finding an objective image quality assessment (IQA) method that
measures the degree of enhancement is very difficult. Actually, 
there may not be "objective" measure
because the aesthetic perception is also individually different.
Anyway, in this paper, we select two IQA metrics which have been adopted
for similar purposes as our case. 
Specifically, we adopt the gradient magnitude similarity deviation (GMSD) 
as a full-reference IQA, and the natural image quality evaluator (NIQE) 
as a no-reference IQA method \cite{ex:gmsd, ex:niqe}. 
The GMSD measures visual distortion between an input image and its enhancement \cite{ex:gmsd}. The lower GMSD means less visual distortion. 
The NIQE is based on the statistical regularities that can be derived from
natural and undistorted images\cite{ex:niqe}. The lower NIQE also means
better quality. 
Table \ref{tb:obj_result} lists these metrics for the
images compared above, where the best and second best numbers
are written in bold. While the proposed algorithm does not always 
show the best performance, it shows quite low values on average. 
Also, there are many cases that both of GMSD and NIQE measures are
lower than others. 

\begin {figure} 
\centering
\includegraphics[width=90mm]{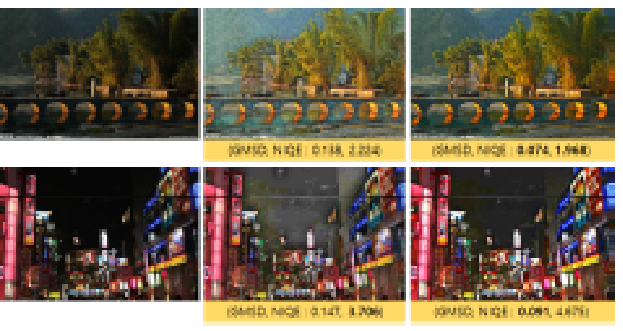}
\caption{Sample images obtained by NPEA and the proposed method; (from left to right in each row) input images, output images by NPEA and ours.}
\label{Figure_Exp_Measure_Error}
\end {figure}

We carried out more experiments on 70 images from 
\cite{npea}, and the average GMSD for NPEA/Ours
is shown to be (0.118/0.065) and NIQE be (3.365/3.469).
That is, our algorithm shows better results in
terms of GMSD, and slightly worse in NIQE than
the NPEA. However, as stated previously, the objective
measure may not be in agreement with the individual's
subjective assessment. As an example, 
Fig.~\ref{Figure_Exp_Measure_Error} shows some
results by NPEA and ours, where it seems
that NIQE prefers stretching the 
brightness to keeping the naturalness or 
suppressing the distortions.
For more subjective comparisons,
refer to large images in our
project page 
(https://github.com/JasonBournePark/EnhanceHDR), 
where materials for our algorithm 
are also available.

\begin{table*}[t]
	\centering
	\caption{GMSD and NIQE measurements.}
	\begin{tabular}{@{} l*{15}{>{$}c<{$}} @{}}
		\toprule
		\multirow{2}{*}{Images} & \multicolumn{2}{c@{}}{$CLAHE$}  & \multicolumn{2}{c@{}}{$CVC$}  & \multicolumn{2}{c@{}}{$MSR$}  & \multicolumn{2}{c@{}}{$FbEM$}  & \multicolumn{2}{c@{}}{$LIME$}  & \multicolumn{2}{c@{}}{$NPEA$}  & \multicolumn{2}{c@{}}{$OURS$} \\
		\cmidrule(l){2-3} \cmidrule(l){4-5} \cmidrule(l){6-7} \cmidrule(l){8-9}  \cmidrule(l){10-11}  \cmidrule(l){12-13} \cmidrule(l){14-15}
		& G & \cellcolor{gray!10}N & G & \cellcolor{gray!10}N & G & \cellcolor{gray!10}N & G & \cellcolor{gray!10}N & G & \cellcolor{gray!10}N & G & \cellcolor{gray!10}N & G & \cellcolor{gray!10}N & \\ \hline
		\midrule
		Tile-roofed house	& 0.134 & \cellcolor{gray!10}1.572 & \textbf{0.045} & \cellcolor{gray!10}1.746 & 0.289 & \cellcolor{gray!10}1.940 & 0.105 & \cellcolor{gray!10}\textbf{1.481} & 0.189 & \cellcolor{gray!10}1.708 & 0.117 & \cellcolor{gray!10}1.557 & \textbf{0.042} & \cellcolor{gray!10}\textbf{1.533} \\
		Wooden ceiling		& 0.130 & \cellcolor{gray!10}1.867 & \textbf{0.022} & \cellcolor{gray!10}1.773 & 0.294 & \cellcolor{gray!10}2.779 & 0.134 & \cellcolor{gray!10}1.826 & 0.223 & \cellcolor{gray!10}2.214 & 0.156 & \cellcolor{gray!10}\textbf{1.526} & \textbf{0.079} & \cellcolor{gray!10}\textbf{1.678} \\
		Daybreak 			& 0.130 & \cellcolor{gray!10}\textbf{3.183} & \textbf{0.069} & \cellcolor{gray!10}3.837 & 0.246 & \cellcolor{gray!10}3.421 & 0.119 & \cellcolor{gray!10}3.483 & 0.192 & \cellcolor{gray!10}\textbf{3.382} & 0.117 & \cellcolor{gray!10}3.637 & \textbf{0.088} & \cellcolor{gray!10}3.631 \\
		Duck 				& 0.152 & \cellcolor{gray!10}\textbf{2.651} & \textbf{0.121} & \cellcolor{gray!10}2.883 & 0.264 & \cellcolor{gray!10}4.246 & 0.155 & \cellcolor{gray!10}2.839 & 0.214 & \cellcolor{gray!10}2.994 & 0.146 & \cellcolor{gray!10}2.813 & \textbf{0.082} & \cellcolor{gray!10}\textbf{2.488} \\
		Entrance 			& 0.148 & \cellcolor{gray!10}3.441 & 0.209 & \cellcolor{gray!10}3.264 & 0.264 & \cellcolor{gray!10}3.780 & \textbf{0.138} & \cellcolor{gray!10}3.524 & 0.221 & \cellcolor{gray!10}3.582 & 0.174 & \cellcolor{gray!10}\textbf{3.160} & \textbf{0.132} & \cellcolor{gray!10}\textbf{3.124} \\ 	\midrule
		Average 			& 0.140 & \cellcolor{gray!10}2.543 & \textbf{0.093} & \cellcolor{gray!10}2.701 & 0.275 & \cellcolor{gray!10}3.233 & 0.130 & \cellcolor{gray!10}2.631 & 0.208 & \cellcolor{gray!10}2.776 & 0.142 & \cellcolor{gray!10}\textbf{2.539} & \textbf{0.085} & \cellcolor{gray!10}\textbf{2.491} \\	\midrule
		\bottomrule
		\label{tb:obj_result}
	\end{tabular}
\end{table*}

\subsection{Discussions}
\subsubsection{Noise reduction}
Generally, expanding the range of an image is accompanied by noise 
amplification especially in dark areas where the signal to
noise ratio is inherently low. Hence our method (SRS) attempts to avoid the
expansion of reflectance in dark areas, and instead increases the 
filtered illumination. Note that the illumination component is
obtained by the WLSF that also reduces the noise. However
the noise is still somewhat amplified,
though less than the case of directly expanding the pixel values.
For reducing the noise amplified by the enhancement, 
we may consider applying a noise reduction algorithm 
as a post-processing step.
However, it should be noted that the noise
reduction accompanies blurring effect, and
filtering the reflectance component may
result in large changes in the pixel values because
the reflectance is described in log scale.
Thus we apply the noise reduction to the illumination 
component only, for keeping the edges and textures
that are in the reflectance component. 

\begin {figure} 
\centering
\includegraphics[width=90mm]{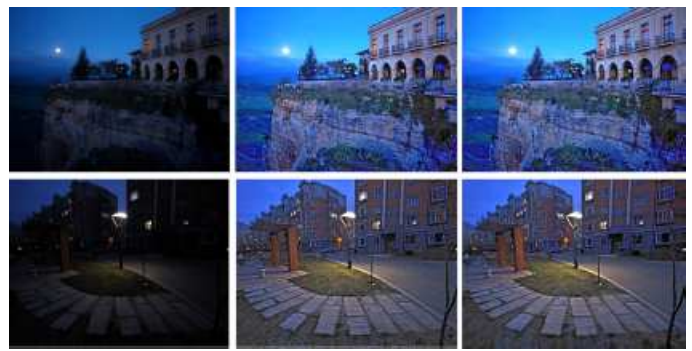}
\caption{Noise reduction results, (from left to right in each row) input image, result of the proposed method and noise suppression result using BM3D.}
\label{Figure13_Exp_Rst_NR}
\end {figure}

As for the noise reduction, recent deep learning based
methods \cite{dncnn,bmcnn} perform better than the
conventional transformation-filtering-based BM3D \cite{ex:bm3d}.
However, since the neural network methods need huge computing power (GPU)
and also since the purpose of adopting the noise reduction in this paper
is just to demonstrate an example of post-processing for further rectification,
we just adopt the BM3D. Fig.~\ref{Figure13_Exp_Rst_NR} shows the
result of this post-processing, where it can be seen that the noise
is considerably reduced without any other side effect such as 
loss of details.

\subsubsection{Detail enhancement on the reflectance}
\begin {figure*} 
\centering
\includegraphics[width=170mm]{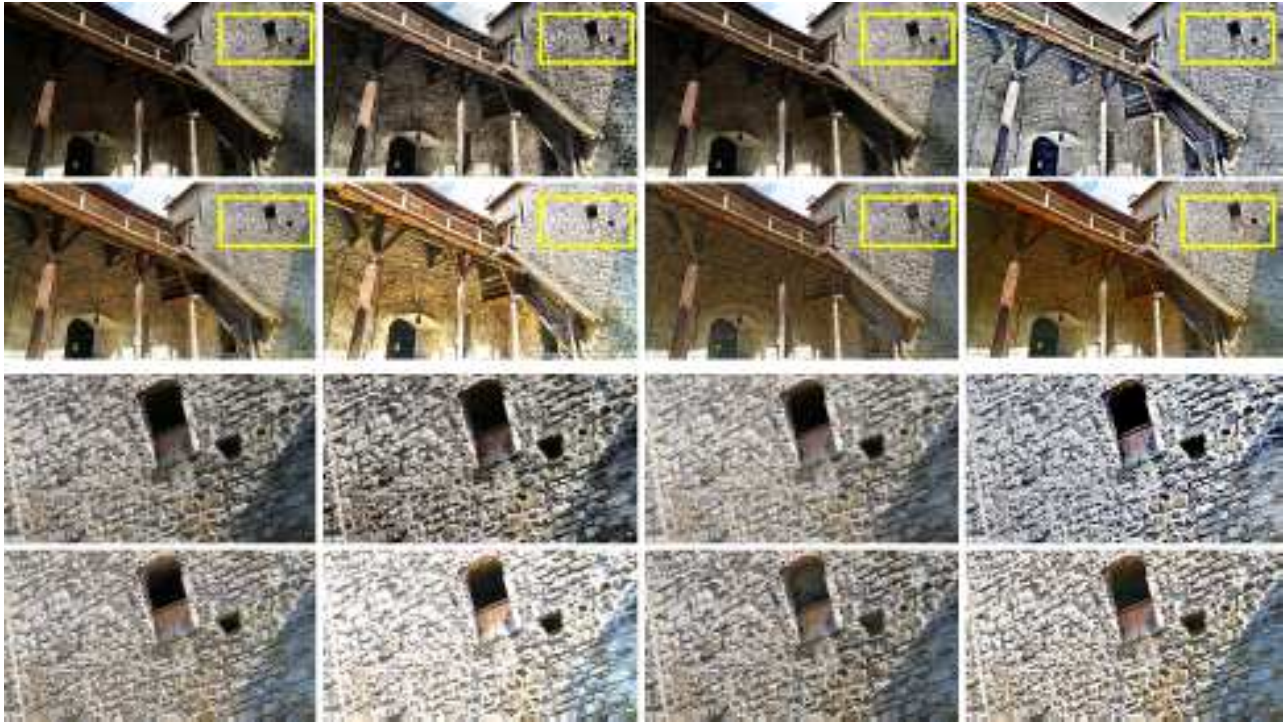}
\caption{Comparison of 'Stone wall' image labled as C25 in 
	MMSPG dataset. From top left to bottom right: input image, CLAHE, 
	CVC, MSR, FbEM, LIME, NPEA and the proposed. The third and fourth
	rows are the magnifications of yellow boxes}
\label{Figure14_Exp_Rst_Details}
\end {figure*}

In this subsection, we discuss the effect of our reflectance manipulation
(SRS in eq.~\ref{eq: tone_repro}) compared to the conventional methods 
that focus on increasing the illumination only. 
In Fig. \ref{Figure14_Exp_Rst_Details}, we present the visual
comparison again for another image, which is a kind of backlight and/or 
non-uniformly illuminated image. Since it has dark shadow and simultaneously 
bright areas, adjusting only the illumination does not improve the 
details in bright areas. In the CLAHE and MSR results, it can be
observed that the details look considerably enhanced but the visual quality 
is degraded. In the result of CVC, only the brightness level is improved. 
The output image is over-enhanced in the case of LIME algorithm and hence 
details in bright areas become less visible. 
The FbEM and NPEA generate relatively well illuminated and comfortable 
images, but they did not much enhance the details in bright areas. 
In the case of proposed method, it appears that both of
bright and dark areas are well enhanced.

\subsubsection{Run-time comparison}
Run-times are compared in Table~\ref{tb:comput_time} for 
several different resolutions. The proposed method requires 
relatively large run-time (denoted as w/ WLSF in the table) 
mainly due to the WLSF for generating the high quality edge-preserved
illumination map \cite{wlsf}. The run-time without the WLSF 
(denoted as w/o WLSF) is much smaller than others. Hence, if
we use a simple illumination estimation method such as 
global smoothing filter (GSF) \cite{fastWLSF}, the run-time (denoted 
as w/ GSF) is less than others though there can be some degradation in
image quality. Specifically, if we use GSF instead of WLSF,
then the GMSD and NIQE measures increase approximately by 
14\% and 2\% respectively. 

\def\HS{\hspace{\fontdimen2\font}} 
\begin{table*}[t]
	\centering
	\caption{Run-time comparison.}
	\begin{tabular}{@{} l*{10}{>{$}c<{$}} @{}}
		\toprule
		\multirow{2}{*}{\parbox{1.2cm}{Image \HS resolution}} & \multirow{2}[3]{*}{$CLAHE$} & \multirow{2}[3]{*}{$CVC$} & \multirow{2}[3]{*}{$MSR$} & \multirow{2}[3]{*}{$FbEM$} & \multirow{2}[3]{*}{$LIME$} & \multirow{2}[3]{*}{$NPEA$} & \multicolumn{3}{c@{}}{$OURS$} \\
		\cmidrule(l){8-10}
		& & & & & & & $w/o \HS WLSF$ & $w/ \HS WLSF$ & $w/ \HS GSF$\\ \hline
		\midrule
		$560 \times 420$ 		& 0.182 & 0.137 & 0.583 & 0.185 & 0.100 & 7.637 & \textbf{0.084} & 0.740  & \textbf{0.095}  \\
		\midrule
		$1038 \times 789$ 		& 1.055 & 0.565 & 1.749 & 1.928 & 1.046 & 27.065 & \textbf{0.267} & 2.537 & \textbf{0.295}  \\
		\midrule
		$1920 \times 1080$ 		& 1.344 & 1.205 & 3.020 & 1.416 & 0.955 & 68.181 & \textbf{0.606} & 6.987 & \textbf{0.783} \\
		\midrule
		$1536 \times 2048$ 		& 2.054 & 2.027 & 4.557 & 2.141 & 1.571 & 103.755 & \textbf{0.916} & 12.684 & \textbf{1.259} \\
		\bottomrule
		\label{tb:comput_time}
	\end{tabular}
\end{table*}

\subsection{HDR image generation and applications}

\subsubsection{Extension to the HDR reverse tone mapper}
The main purpose of this paper is the image enhancement and so we
have focused on evaluating our algorithm in comparison with
other enhancement methods. In addition to this, since our method
is basically a single image HDR, we briefly test it as an 
HDR imaging method. 

\begin {figure*} 
\centering
\includegraphics[width=170mm]{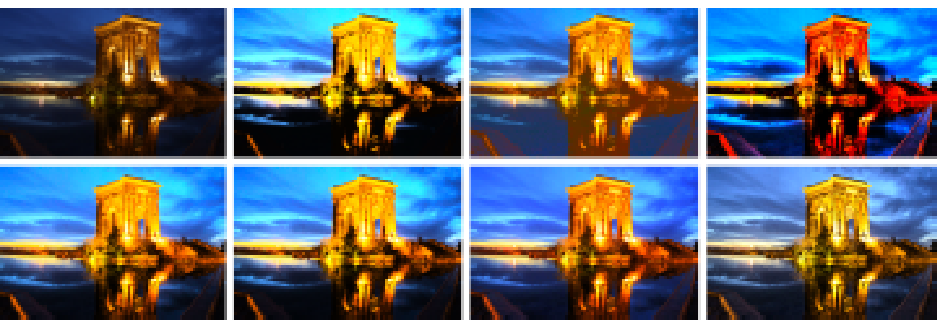}
\caption{Tone mapped images of HDR images generated by several rTMOs. From top 
	left to bottom right: input LDR image, Akyuz et. al's \cite{ex:itmo-akyuz}, 
	Banterle et. al's \cite{ex:itmo-ban}, Huo et. al's \cite{ex:itmo-huo}, 
	Meylan et. al's \cite{ex:itmo-meylan}, Rempel et. al's in \cite{ex:itmo-rempel}, 
	Wang et. al's \cite{hdr-pexp-fusion} and the proposed method.}
\label{Figure15_Exp_Rst_iTMOs}
\end {figure*}

Generating an HDR image from a single exposure image is called a single
image HDR in some software tools or reverse tone mapping 
\cite{ex:itmo-banterle, ex:itmo-rempel, 
	hdr-pexp-fusion, ex:itmo-color, ex:hdr-toolbox}.
The reverse tone mapping operation (rTMO) is an ill-posed problem because it tries
to find the information that was lost during the capturing process.
In this subsection, we test whether the pseudo multi-exposures generated
by our algorithm can be used for generating an HDR image, i.e., we test
the feasibility of our method for rTMO. For this, we generate some of the 
HDR images by the reverse tone mapping functions provided in the HDR Matlab 
toolbox \cite{ex:hdr-toolbox} and also provided by the author in 
\cite{hdr-pexp-fusion}. We set the parameters of our algorithm the same as
in the image enhancement experiments above. The only additional processing is to 
generate an RGBE image (.hdr format) by encoding the enhanced luminance 
information ($L'$) based on the target luminance range of a display. 
The parameters in each of compared rTMOs are set to be the default values, 
and the minimum and maximum luminance of target display are set to 
0.0015 $cd/m^2$ (nits) and 3000 $cd/m^2$ respectively. 
The comparison is shown in Fig.~\ref{Figure15_Exp_Rst_iTMOs}, where the 
global tone-mapping function proposed in \cite{pa:reinhard-tone-repro} is used 
for the LDR display. It can be seen that 
the proposed algorithm is comparable to the other conventional rTMOs. 

\subsubsection{Pre-processing and post-processing}
\begin {figure} 
\centering
\includegraphics[width=80mm]{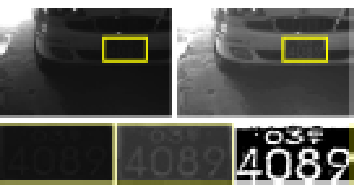}
\caption{Pre-processing for ANPR. 
	Top left and right: input image and enhanced image by our proposed algorithm. 
	Bottom, from left to right: enlargement of original, enhanced, binarized number plate.}
\label{Figure16_Exp_Rst_ANPR}
\end {figure}

We demonstrate that the proposed method is also suitable as a
pre-processing step for the computer vision tasks. For example,
we show that it is very effective for enhancing the objects
in low illumination conditions such that it can enhance the
performance of automatic number plate recognition (ANPR) system \cite{ex:anpr}. 
Specifically, the number plate in the left image of Fig.~\ref{Figure16_Exp_Rst_ANPR} 
is not detected, but after applying our algorithm we obtain the image 
as shown in the right side. Then we can successfully detect the plate and
binarize it as shown in the bottom figures.

\begin {figure*} 
\centering
\includegraphics[width=170mm]{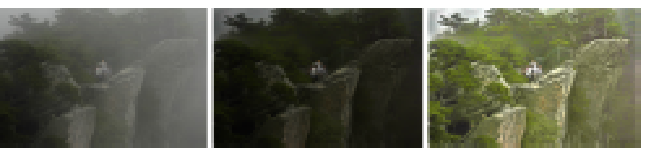}
\caption{Post-processing of haze removed image. From left to right: 
	input image, haze-removed image and enhanced image by our proposed algorithm.}
\label{Figure17_Exp_Rst_DCP}
\end {figure*}

The proposed method can also be used as a post-processing step for further
enhancing the images that are already processed for other purposes. For example, 
let us consider our method as a post-processor for haze removed images. 
Fig.~\ref{Figure17_Exp_Rst_DCP} shows an input image with
the haze (left), its enhancement by using the dark channel prior (DCP) method
in \cite{ex:DCP} (center), and its enhancement by our algorithm (right). 
In general, the results of dehazing methods suffer from losing details and 
lack of brightness, and our algorithm can enhance the haze-removed images
to be more visually pleasing. 

\section{Conclusion}
\label{s:Conclusion}
We have proposed an image enhancement algorithm based on the
single image HDR. The input image is first
decomposed into reflectance and illumination components based
on the Retinex theory.
The reflectance image is separately manipulated to enhance
the details in bright areas, and the illumination component is
scaled up and down to generate several virtual illuminations that
correspond to certain exposure values. That is, we generate
several multi-exposure illuminations from the single input, and 
they are fused to be an enhanced illumination. Then we multiply the 
enhanced reflectance and fused illumination to obtain the final enhanced 
image. Experiments show that the proposed method produces visually
pleasing results, and also yields comparable or better results than
the conventional methods in terms of some objective measures.



\section*{Acknowledgment}
This study is conducted in the first author's academic training course 
supported by Samsung Electronics Co., Ltd.

\ifCLASSOPTIONcaptionsoff
  \newpage
\fi



%
\bibliographystyle{IEEEtran}
\bibliography{IEEEabrv,IEEEexample}

\begin{thebibliography}{10}
\providecommand{\url}[1]{#1}
\csname url@samestyle\endcsname
\providecommand{\newblock}{\relax}
\providecommand{\bibinfo}[2]{#2}
\providecommand{\BIBentrySTDinterwordspacing}{\spaceskip=0pt\relax}
\providecommand{\BIBentryALTinterwordstretchfactor}{4}
\providecommand{\BIBentryALTinterwordspacing}{\spaceskip=\fontdimen2\font plus
\BIBentryALTinterwordstretchfactor\fontdimen3\font minus
  \fontdimen4\font\relax}
\providecommand{\BIBforeignlanguage}[2]{{%
\expandafter\ifx\csname l@#1\endcsname\relax
\typeout{** WARNING: IEEEtran.bst: No hyphenation pattern has been}%
\typeout{** loaded for the language `#1'. Using the pattern for}%
\typeout{** the default language instead.}%
\else
\language=\csname l@#1\endcsname
\fi
#2}}
\providecommand{\BIBdecl}{\relax}
\BIBdecl

\bibitem{gonzalez}
R.~C. Gonzalez and R.~E. Woods, ``Image processing,'' \emph{Digital image
  processing}, vol.~2, 2007.

\bibitem{baxes}
G.~A. Baxes, \emph{Digital image processing: principles and
  applications}.\hskip 1em plus 0.5em minus 0.4em\relax Wiley New York, 1994.

\bibitem{ldr}
C.~Lee, C.~Lee, and C.-S. Kim, ``Contrast enhancement based on layered
  difference representation of 2d histograms,'' \emph{IEEE Transactions on
  Image Processing}, vol.~22, no.~12, pp. 5372--5384, 2013.

\bibitem{gum}
G.~Deng, ``A generalized unsharp masking algorithm,'' \emph{IEEE transactions
  on Image Processing}, vol.~20, no.~5, pp. 1249--1261, 2011.

\bibitem{febm}
X.~Fu, D.~Zeng, Y.~Huang, Y.~Liao, X.~Ding, and J.~Paisley, ``A fusion-based
  enhancing method for weakly illuminated images,'' \emph{Signal Processing},
  vol. 129, pp. 82--96, 2016.

\bibitem{npea}
S.~Wang, J.~Zheng, H.-M. Hu, and B.~Li, ``Naturalness preserved enhancement
  algorithm for non-uniform illumination images,'' \emph{IEEE Transactions on
  Image Processing}, vol.~22, no.~9, pp. 3538--3548, 2013.

\bibitem{lime}
X.~Guo, Y.~Li, and H.~Ling, ``Lime: Low-light image enhancement via
  illumination map estimation,'' \emph{IEEE Transactions on Image Processing},
  2016.

\bibitem{ahe}
S.~M. Pizer, E.~P. Amburn, J.~D. Austin, R.~Cromartie, A.~Geselowitz, T.~Greer,
  B.~ter Haar~Romeny, J.~B. Zimmerman, and K.~Zuiderveld, ``Adaptive histogram
  equalization and its variations,'' \emph{Computer vision, graphics, and image
  processing}, vol.~39, no.~3, pp. 355--368, 1987.

\bibitem{bi-HE}
D.-I. Yoon, ``Contrast enhancement using brightness preserving bi-histogram
  equalization,'' \emph{School of Information and Computer Eng.}, 2008.

\bibitem{clahe}
E.~D. Pisano, S.~Zong, B.~M. Hemminger, M.~DeLuca, R.~E. Johnston, K.~Muller,
  M.~P. Braeuning, and S.~M. Pizer, ``Contrast limited adaptive histogram
  equalization image processing to improve the detection of simulated
  spiculations in dense mammograms,'' \emph{Journal of Digital imaging},
  vol.~11, no.~4, pp. 193--200, 1998.

\bibitem{hist-modi-framework}
T.~Arici, S.~Dikbas, and Y.~Altunbasak, ``A histogram modification framework
  and its application for image contrast enhancement,'' \emph{IEEE Transactions
  on image processing}, vol.~18, no.~9, pp. 1921--1935, 2009.

\bibitem{wlsf}
Z.~Farbman, R.~Fattal, D.~Lischinski, and R.~Szeliski, ``Edge-preserving
  decompositions for multi-scale tone and detail manipulation,'' in \emph{ACM
  Transactions on Graphics (TOG)}, vol.~27, no.~3.\hskip 1em plus 0.5em minus
  0.4em\relax ACM, 2008, p.~67.

\bibitem{per-pix-exp}
E.~P. Bennett and L.~McMillan, ``Video enhancement using per-pixel virtual
  exposures,'' in \emph{ACM Transactions on Graphics (TOG)}, vol.~24,
  no.~3.\hskip 1em plus 0.5em minus 0.4em\relax ACM, 2005, pp. 845--852.

\bibitem{fbf-hdr}
F.~Durand and J.~Dorsey, ``Fast bilateral filtering for the display of
  high-dynamic-range images,'' in \emph{ACM transactions on graphics (TOG)},
  vol.~21, no.~3.\hskip 1em plus 0.5em minus 0.4em\relax ACM, 2002, pp.
  257--266.

\bibitem{auto-exp-correct}
Y.~Wang, S.~Zhuo, D.~Tao, J.~Bu, and N.~Li, ``Automatic local exposure
  correction using bright channel prior for under-exposed images,''
  \emph{Signal Processing}, vol.~93, no.~11, pp. 3227--3238, 2013.

\bibitem{flash-photo}
E.~Eisemann and F.~Durand, ``Flash photography enhancement via intrinsic
  relighting,'' \emph{ACM transactions on graphics (TOG)}, vol.~23, no.~3, pp.
  673--678, 2004.

\bibitem{cent-surr-retinex}
D.~J. Jobson, Z.-u. Rahman, and G.~A. Woodell, ``Properties and performance of
  a center/surround retinex,'' \emph{IEEE transactions on image processing},
  vol.~6, no.~3, pp. 451--462, 1997.

\bibitem{meylan}
L.~Meylan and S.~S{\"u}sstrunk, ``Color image enhancement using a retinex-based
  adaptive filter,'' in \emph{Conference on Colour in Graphics, Imaging, and
  Vision}, vol. 2004, no.~1.\hskip 1em plus 0.5em minus 0.4em\relax Society for
  Imaging Science and Technology, 2004, pp. 359--363.

\bibitem{framework-retinex}
R.~Kimmel, M.~Elad, D.~Shaked, R.~Keshet, and I.~Sobel, ``A variational
  framework for retinex,'' \emph{International Journal of computer vision},
  vol.~52, no.~1, pp. 7--23, 2003.

\bibitem{hdr-vexp-fusion}
C.-H. Lee, L.-H. Chen, and W.-K. Wang, ``Image contrast enhancement using
  classified virtual exposure image fusion,'' \emph{IEEE Transactions on
  Consumer Electronics}, vol.~58, no.~4, 2012.

\bibitem{hdr-pexp-fusion}
T.-H. Wang, C.-W. Chiu, W.-C. Wu, J.-W. Wang, C.-Y. Lin, C.-T. Chiu, and J.-J.
  Liou, ``Pseudo-multiple-exposure-based tone fusion with local region
  adjustment,'' \emph{IEEE Transactions on Multimedia}, vol.~17, no.~4, pp.
  470--484, 2015.

\bibitem{hdr-reinhard}
E.~Reinhard, W.~Heidrich, P.~Debevec, S.~Pattanaik, G.~Ward, and K.~Myszkowski,
  \emph{High dynamic range imaging: acquisition, display, and image-based
  lighting}.\hskip 1em plus 0.5em minus 0.4em\relax Morgan Kaufmann, 2010.

\bibitem{hdr-exp-fusion}
T.~Mertens, J.~Kautz, and F.~Van~Reeth, ``Exposure fusion: A simple and
  practical alternative to high dynamic range photography,'' in \emph{Computer
  Graphics Forum}, vol.~28, no.~1.\hskip 1em plus 0.5em minus 0.4em\relax Wiley
  Online Library, 2009, pp. 161--171.

\bibitem{an1}
J.~An, S.~H. Lee, J.~G. Kuk, and N.~I. Cho, ``A multi-exposure image fusion
  algorithm without ghost effect,'' in \emph{Acoustics, Speech and Signal
  Processing (ICASSP), 2011 IEEE International Conference on}.\hskip 1em plus
  0.5em minus 0.4em\relax IEEE, 2011, pp. 1565--1568.

\bibitem{an2}
J.~An, S.~J. Ha, and N.~I. Cho, ``Probabilistic motion pixel detection for the
  reduction of ghost artifacts in high dynamic range images from multiple
  exposures,'' vol. 2014, no.~1.\hskip 1em plus 0.5em minus 0.4em\relax
  EURASIP, 2014, pp. 1--15.

\bibitem{an3}
J.~\vspace{0mm} An, S.~J. Ha, and N.~I. Cho, ``Reduction of ghost effect in
  exposure fusion by detecting the ghost pixels in saturated and non-saturated
  regions,'' in \emph{Acoustics, Speech and Signal Processing (ICASSP), 2012
  IEEE International Conference on}.\hskip 1em plus 0.5em minus 0.4em\relax
  IEEE, 2012, pp. 1101--1104.

\bibitem{hdr-sensor-array}
S.~K. Nayar and T.~Mitsunaga, ``High dynamic range imaging: Spatially varying
  pixel exposures,'' in \emph{Computer Vision and Pattern Recognition, 2000.
  Proceedings. IEEE Conference on}, vol.~1.\hskip 1em plus 0.5em minus
  0.4em\relax IEEE, 2000, pp. 472--479.

\bibitem{cvc}
T.~Celik and T.~Tjahjadi, ``Contextual and variational contrast enhancement,''
  \emph{IEEE Transactions on Image Processing}, vol.~20, no.~12, pp.
  3431--3441, 2011.

\bibitem{ex:DCP}
K.~He, J.~Sun, and X.~Tang, ``Single image haze removal using dark channel
  prior,'' \emph{IEEE transactions on pattern analysis and machine
  intelligence}, vol.~33, no.~12, pp. 2341--2353, 2011.

\bibitem{pa:retinex-wlsf}
E.~Zhang, H.~Yang, M.~Xu \emph{et~al.}, ``A novel tone mapping method for high
  dynamic range image by incorporating edge-preserving filter into method based
  on retinex,'' \emph{Applied Mathematics \& Information Sciences}, vol.~9,
  no.~1, pp. 411--417, 2015.

\bibitem{pa:de}
D.-Y. Tsai, Y.~Lee, and E.~Matsuyama, ``Information entropy measure for
  evaluation of image quality,'' \emph{Journal of digital imaging}, vol.~21,
  no.~3, pp. 338--347, 2008.

\bibitem{pa:hdrsoft-web}
\BIBentryALTinterwordspacing
The photomatix website. [Online]. Available:
  \url{https://www.hdrsoft.com/index.html/}
\BIBentrySTDinterwordspacing

\bibitem{pa:adapt-exp-correct}
G.~Messina, A.~Castorina, S.~Battiato, and A.~Bosco, ``Image quality
  improvement by adaptive exposure correction techniques,'' in \emph{Multimedia
  and Expo, 2003. ICME'03. Proceedings. 2003 International Conference on},
  vol.~1.\hskip 1em plus 0.5em minus 0.4em\relax IEEE, 2003, pp. I--549.

\bibitem{pa:kim-tone-repro}
K.~Kim, J.~Bae, and J.~Kim, ``Natural hdr image tone mapping based on
  retinex,'' \emph{IEEE Transactions on Consumer Electronics}, vol.~57, no.~4,
  2011.

\bibitem{ex:mmspg}
P.~Korshunov, H.~Nemoto, A.~Skodras, and T.~Ebrahimi, ``Crowdsourcing-based
  evaluation of privacy in hdr images,'' in \emph{SPIE Photonics Europe}.\hskip
  1em plus 0.5em minus 0.4em\relax International Society for Optics and
  Photonics, 2014, pp. 913\,802--913\,802.

\bibitem{ex:gmsd}
W.~Xue, L.~Zhang, X.~Mou, and A.~C. Bovik, ``Gradient magnitude similarity
  deviation: A highly efficient perceptual image quality index,'' \emph{IEEE
  Transactions on Image Processing}, vol.~23, no.~2, pp. 684--695, 2014.

\bibitem{ex:niqe}
A.~Mittal, R.~Soundararajan, and A.~C. Bovik, ``Making a 'completely blind'
  image quality analyzer,'' \emph{IEEE Signal Processing Letters}, vol.~20,
  no.~3, pp. 209--212, 2013.

\bibitem{dncnn}
K.~Zhang, W.~Zuo, Y.~Chen, D.~Meng, and L.~Zhang, ``Beyond a gaussian denoiser:
  Residual learning of deep cnn for image denoising,'' \emph{IEEE Transactions
  on image Processing}, 2017.

\bibitem{bmcnn}
B.~Ahn and N.~I. Cho, ``Block-matching convolutional neural network for image
  denoising,'' \emph{arXiv preprint arXiv:1704.00524}, 2017.

\bibitem{ex:bm3d}
K.~Dabov, A.~Foi, V.~Katkovnik, and K.~Egiazarian, ``Image denoising by sparse
  3-d transform-domain collaborative filtering,'' \emph{IEEE Transactions on
  image processing}, vol.~16, no.~8, pp. 2080--2095, 2007.

\bibitem{fastWLSF}
D.~Min, S.~Choi, J.~Lu, B.~Ham, K.~Sohn, and M.~N. Do, ``Fast global image
  smoothing based on weighted least squares,'' \emph{IEEE Transactions on Image
  Processing}, vol.~23, no.~12, pp. 5638--5653, 2014.

\bibitem{ex:itmo-banterle}
F.~Banterle, P.~Ledda, K.~Debattista, and A.~Chalmers, ``Inverse tone
  mapping,'' in \emph{Proceedings of the 4th international conference on
  Computer graphics and interactive techniques in Australasia and Southeast
  Asia}.\hskip 1em plus 0.5em minus 0.4em\relax ACM, 2006, pp. 349--356.

\bibitem{ex:itmo-rempel}
A.~G. Rempel, M.~Trentacoste, H.~Seetzen, H.~D. Young, W.~Heidrich,
  L.~Whitehead, and G.~Ward, ``Ldr2hdr: on-the-fly reverse tone mapping of
  legacy video and photographs,'' in \emph{ACM transactions on graphics (TOG)},
  vol.~26, no.~3.\hskip 1em plus 0.5em minus 0.4em\relax ACM, 2007, p.~39.

\bibitem{ex:itmo-color}
C.-R. Chen, C.-T. Chiu, and Y.-C. Chang, ``Inverse tone mapping operator
  evaluation using blind image quality assessment,'' in \emph{Asia--Pacific
  Sign. and Information Proc. Association Annual Summit and Conf., APSIPA Oct},
  2011.

\bibitem{ex:hdr-toolbox}
F.~Banterle, A.~Artusi, K.~Debattista, and A.~Chalmers, \emph{Advanced high
  dynamic range imaging: theory and practice}.\hskip 1em plus 0.5em minus
  0.4em\relax CRC press, 2011.

\bibitem{pa:reinhard-tone-repro}
E.~Reinhard, M.~Stark, P.~Shirley, and J.~Ferwerda, ``Photographic tone
  reproduction for digital images,'' \emph{ACM transactions on graphics (TOG)},
  vol.~21, no.~3, pp. 267--276, 2002.

\bibitem{ex:itmo-akyuz}
A.~O. Aky{\"u}z, R.~Fleming, B.~E. Riecke, E.~Reinhard, and H.~H. B{\"u}lthoff,
  ``Do hdr displays support ldr content?: a psychophysical evaluation,''
  \emph{ACM Transactions on Graphics (TOG)}, vol.~26, no.~3, p.~38, 2007.

\bibitem{ex:itmo-ban}
F.~Banterle, P.~Ledda, K.~Debattista, and A.~Chalmers, ``Expanding low dynamic
  range videos for high dynamic range applications,'' in \emph{Proceedings of
  the 24th Spring Conference on Computer Graphics}.\hskip 1em plus 0.5em minus
  0.4em\relax ACM, 2008, pp. 33--41.

\bibitem{ex:itmo-huo}
Y.~Huo, F.~Yang, and V.~Brost, ``Dodging and burning inspired inverse tone
  mapping algorithm,'' \emph{J. Comput. Inf. Syst.}, vol.~9, no.~9, pp.
  3461--3468, 2013.

\bibitem{ex:itmo-meylan}
L.~Meylan, S.~Daly, and S.~S{\"u}sstrunk, ``The reproduction of specular
  highlights on high dynamic range displays,'' in \emph{Color and Imaging
  Conference}, vol. 2006, no.~1.\hskip 1em plus 0.5em minus 0.4em\relax Society
  for Imaging Science and Technology, 2006, pp. 333--338.

\bibitem{ex:anpr}
R.~Lotufo, A.~Morgan, and A.~Johnson, ``Automatic number-plate recognition,''
  in \emph{Image Analysis for Transport Applications, IEE Colloquium on}.\hskip
  1em plus 0.5em minus 0.4em\relax IET, 1990, pp. 6--1.

\end{thebibliography}


%

\end{document}